\documentclass[]{article}
\usepackage[round]{natbib}
\usepackage[final]{nips_2017}
\usepackage[utf8]{inputenc}
\usepackage[T1]{fontenc}
\usepackage{hyperref}
\usepackage{url}
\usepackage{booktabs}
\usepackage{amsfonts}
\usepackage{nicefrac}
\usepackage{microtype}

\newcommand{\INDSTATE}[1][1]{\STATE\hspace{#1\algorithmicindent}}

\usepackage{amsmath}
\usepackage{amsfonts}
\usepackage{algorithm}
\usepackage{algorithmic}
\usepackage{wrapfig}
\usepackage{graphicx}
\usepackage{float}
\usepackage{graphicx}
\usepackage{caption}
\usepackage{subcaption}

\title{TIDBD: Adapting Temporal-difference Step-sizes Through Stochastic Meta-descent}

\author{
  Alex Kearney\thanks{All authors are with the Reinforcement Learning and Artificial Intelligence Laboratory (RLAI).}\\
  Department of Computing Science, University of Alberta \\
  Edmonton, AB, Canada, T6G 2E1\\
  \texttt{kearney@ualberta.ca} \\
  \And
  Vivek Veeriah \\
  Department of Computing Science, University of Alberta \\
  Edmonton, AB, Canada, T6G 2E1\\
  \texttt{vivekveeriah@ualberta.ca} \\
  \AND
  Jaden B. Travnik \\
  Department of Computing Science, University of Alberta \\
  Edmonton, AB, Canada, T6G 2E1\\
  \texttt{travnik@ualberta.ca} \\
  \And
  Richard S. Sutton \\
  Department of Computing Science, University of Alberta \\
  Edmonton, AB, Canada, T6G 2E1\\
  \texttt{rsutton@ualberta.ca} \\
  \And
  Patrick M. Pilarski \\
  Departments of Computing Science \& Medicine, University of Alberta \\
  Edmonton, AB, Canada, T6G 2E1\\
  \texttt{pilarski@ualberta.ca} \\
}

\begin{document}

\maketitle
\begin{abstract}

In this paper, we introduce a method for adapting the step-sizes of temporal difference (TD) learning. The performance of TD methods often depends on well chosen step-sizes, yet few algorithms have been developed for setting the step-size automatically for TD learning. An important limitation of current methods is that they adapt a single step-size shared by all the weights of the learning system. A vector step-size enables greater optimization by specifying parameters on a per-feature basis. Furthermore, adapting parameters at different rates has the added benefit of being a simple form of representation learning. We generalize Incremental Delta Bar Delta (IDBD)---a vectorized adaptive step-size method for supervised learning---to TD learning, which we name TIDBD. We demonstrate that TIDBD is able to find appropriate step-sizes in both stationary and non-stationary prediction tasks, outperforming ordinary TD methods and TD methods with scalar step-size adaptation; we demonstrate that it can differentiate between features which are relevant and irrelevant for a given task, performing representation learning; and we show on a real-world robot prediction task that TIDBD is able to outperform ordinary TD methods and TD methods augmented with AlphaBound and RMSprop.

\end{abstract}
\section{Step-size adaptation in temporal-difference learning}

The problem of how to set step-sizes automatically is an important one for machine learning. The performance of many learning methods depends on a step-size parameter that scales weight updates. To ease the burden on practitioners, it is desirable to set this step-size parameter algorithmically, and over the years many such methods have been proposed. Some of these are fixed schedules, but in principle it is better to adapt the step-size based on experience. Such \emph{step-size adaptation} methods are more suitable for large and long-lived machine learning systems, and are the focus of the present work.

Several interesting issues have been explored within step-size adaptation research. One issue is whether there is a single global step-size shared by all the weights in the learning system, or whether each weight has its own step size. The former is simpler of course, but the latter can be more powerful \citep{sutton1992adapting,schraudolph1999local}. Another issue is whether step-sizes always decrease, or whether they can increase as well as decrease over time. While strictly decreasing step-sizes are often effective in stationary problems, in non-stationary problems it is sometimes advantageous to increase the step-size.

In this paper, we focus on step-size adaptation in Temporal Difference (TD) learning methods. TD methods form key components of many reinforcement learning algorithms. Step-size adaptation in TD learning has received relatively little attention and involves interesting challenges that are less pressing in other learning methods. In particular, TD learning uses learned estimates as targets for further learning; this is known as \emph{bootstrapping}, and because of it TD learning always involves a degree of nonstationarity, which, as noted above, increases the need for step-sizes to increase as well as decrease. 

In this paper, we seek a step-size adaptation method for TD learning that satisfies the two criteria discussed above: 1) step-sizes should be able to both increase and decrease in order to compensate for the non-stationarity in both TD learning, and 2) there should be a vector of many step-sizes in order to specify the step-size on a per-feature basis. 

None of the existing methods for step-size adaptation in TD learning satisfy both of our criteria while also performing well in practice. HL($\lambda$) \citep{hutter2007temporal} and AlphaBound \citep{dabney2012adaptive} have a single step-size which only decreases in value. RMSprop \citep{tieleman2012lecture} satisfies our criteria and can be trivially generalized to TD, however, it does not perform well in TD Learning---as we demonstrate in this paper.

The work which has come closest to satisfying our criteria is the work by \cite{dabney2014adaptive} on SID and NOSID. These methods have increasing and decreasing step-sizes and work well within TD learning, but only adapt a single global step-size. These methods are based on IDBD---a method for supervised learning which adapts it's step-size through stochastic meta-descent\citep{sutton1992adapting}. We adopt a similar approach in this paper. Like \cite{dabney2014adaptive}, we extend IDBD to TD, but unlike in his work, we retain IDBD's vectorized step-size adaptation. We name our method TIDBD (a shortened version of TD-IDBD, pronounced tid-bid). By generalizing IDBD to TD with vectorized step-sizes, TIDBD meets our criteria for a step-size adaptation method for TD learning. 

\section{Markov reward processes}

Prediction problems are of importance to many real-world applications. Being able to anticipate the value of signal from the environment can be seen to be knowledge which is acquired through interaction \citep{sutton2011horde}. For example, being able to anticipate the signal of a servomotor can inform a robot's decisions and form a key component of their control systems \citep{edwards2016application, sherstan2015collaborative}: predictions are a way for machines to anticipate events in their world and inform their decision-making processes. 

Predictions may be thought of as estimating the value of a state in a Markov Reward Process (MRP). A MRP is described by the tuple $ < S,p,r,\gamma> $ where $ S $ is the set of all states, $ p(s^{\prime} | s) $ describes the probability of a transition from a state $ s \in S $ to a new state $ s^{\prime} \in S $, $ r(s,s^{\prime}) $ describes the reward observed on a transition from $s$ to $s^{\prime}$, and $0 \le \gamma \le 1$ is a discount factor which determines how future reward is weighted. The goal in an MRP is to learn a value function $V(s)$ which estimates the expected return from a given state $v^{*}(s):= \mathbb{E} \{ G_t | S_t = s \}$, where the return is $G_t := \sum^{\infty}_{i=1} \gamma^{i-1} R_{t+i}$, or the discounted sum of all future rewards. Within the context of an MRP, a prediction is an estimation of the value of a state---the discounted future return of a signal. For example, the prediction signal could be the position of a robot's gripper, which is later used as an input into a robot's control system.

\section{TIDBD}

We introduce Temporal-Difference Incremental Delta Bar Delta (TIDBD), an extension of IDBD to TD learning. First, we discuss TIDBD's algorithmic implementation (Algorithm \ref{TIDBD_alg}); a derivation as stochastic meta-descent is in the following section. As with conventional TD learning methods, TIDBD's task is to learn an estimate $ V (s | \boldsymbol{w}) $ of the true value of a state $v^*(s)$, where $ \boldsymbol{w} \in \mathbb{R}^{n} $ is the weight vector. TIDBD estimates the value of a state using linear function approximation $V(s | \boldsymbol{w}) = \boldsymbol{w}^\top \phi(s)$: the linear combination of a vector of learned weights $\boldsymbol{w}$ and a feature vector $\phi(s)$ approximating the state $s \in S$. The weights $\boldsymbol{w}$ are updated as follows by moving them in the direction of the TD error $\delta$ by the step-size $\alpha$, where $0 < \alpha$. Note that unlike other TD methods, TIDBD's $\alpha$ is specified for a specific time-step $t$, as its value changes over time.

\begin{algorithm}[h]
\caption{TIDBD($\lambda$)}
\begin{algorithmic}[1]
\STATE Initialize vectors $\boldsymbol{H} \in {0}^{n}$, $\boldsymbol{z} \in {0}^{n} $, and both $\boldsymbol{w} \in \mathbb{R}^{n}$ and $\boldsymbol{\beta} \in \mathbb{R}^{n}$ arbitrarily; initialize a scalar $\theta$; observe state $S$
\STATE Repeat for each observation $s^{\prime}$ and reward $R$: 
  \INDSTATE[1] $\delta \gets R + \gamma \boldsymbol{w}^\top \phi(s^{\prime}) - \boldsymbol{w}^\top \phi(s)$
  \INDSTATE[1] For element $i = 1, 2, \cdots, n $: 
  	\INDSTATE[2] $\beta_i \gets \beta_i + \theta \delta \phi_i(s) H_i$
    \INDSTATE[2] $\alpha_i \gets e^{\beta_i}$
  	\INDSTATE[2] $z_i \gets z_i  \gamma \lambda + \phi_i(s)$
  	\INDSTATE[2] $w_i \gets w_i + \alpha_i \delta z_i$
  	\INDSTATE[2] $H_i \gets H_i [1-\alpha_i\phi_i(s) z_i]^+ + \alpha_i\delta z_i$ 
  \INDSTATE[1] $s \gets s^{\prime}$
\end{algorithmic}
\label{TIDBD_alg}
\end{algorithm}

Eligibility traces, denoted $\boldsymbol{z}$ on line 7, allow the currently observed reward to be attributed to previous experiences; eligibility traces can be thought of as a decaying history of visited states. We make the distinction between two types of traces: accumulating traces (shown in Alg. \ref{TIDBD_alg}), which simply continue adding to $\boldsymbol{z}$ for each visit; and replacing traces, which will replace the value $\boldsymbol{z}$ for each visit to a state. The former can cause instability, as features can have a weight greater than 1 given to their updates as a result of multiple visits to the same state. The TD error $\delta$ on Line 3 is the difference between the predicted return for the current state $V(s | \boldsymbol{w})$ and the estimate of future return $R + \gamma V(s^{\prime} | \boldsymbol{w})$. For a more detailed explanation on TD learning and eligibility traces, please consult the description given by \cite{sutton1998reinforcement}.

TIDBD adds to ordinary TD by updating the values of the step-size $\alpha$ on per-time-step basis; we define our step-size vector using an exponentiation of the $\boldsymbol{\beta}$ vector (Line 6). By exponentiating the step-size parameters, we ensure that the step-size is always positive and that the step-size moves by geometric steps. No fixed meta step-size would appropriate for all features; moving each feature's step-size by a percentage of it's current value is beneficial.

The vector $\boldsymbol{\beta}$ is the set of parameters we adapt in order to change the step-size vector  $\boldsymbol{\alpha}$. On line 5, a meta step-size parameter $\theta$ is used to move $\boldsymbol{\beta}$ at every time-step; $\boldsymbol{\beta}$ is updated in the direction of the most recent change to the weight vector, $\delta \phi(s)$, as scaled by the elements of $\boldsymbol{H}$. The vector $\boldsymbol{H}$ is a decaying trace of the current updates to the weight vector via $\delta \phi(s)$ and previous updates (Line 9). Here $[x]^+$ is $x$ for $x>0$ and $0$ for all other values of $x$. 

This has the effect of changing $\boldsymbol{\beta}$ based on what can be considered the correlation between current updates to our weight vector and previous updates. If the current weight update $\delta \phi(s)$ is strongly correlated with previous weight updates $\boldsymbol{H}$, then we are making many updates in the same direction and it would have been a more efficient use of data to have made a larger update instead. If the updates are negatively correlated, then we have overshot the target value, and should have used smaller step-sizes and smaller updates.

\section{TIDBD derivation}

We now derive TIDBD as stochastic meta-descent. We start the derivation of TIDBD by describing the update rule for $\boldsymbol{\beta}$---the weights with which we define our step-size.

\begin{equation} \label{betaintro}
\begin{split}
\beta_i(t+1) & = \beta_i(t) - \frac{1}{2} \theta \frac{\partial \delta^2(t)}{\partial \beta_i} \\
\end{split}
\end{equation}

TIDBD learns it's step-size parameters $\boldsymbol{
\beta}$, by moving them in the direction of the meta-gradient $\frac{\partial \delta^2(t)}{\partial \beta_i}$. Here, our meta step-size is $-\frac{1}{2}\theta$. In (\ref{beta approx}), we simplify the $\boldsymbol{\beta}$ update by approximating $ \sum_j \frac{\partial \delta^2(t)}{\partial \boldsymbol{w}_j(t)} \frac{\delta \boldsymbol{w}_j(t)}{\delta \beta_i} $ as $\frac{\partial \boldsymbol{w}_j(t)}{\partial \beta_i} \approx 0$ where $i \neq j$. We do this because the effect of changing the step-size for a particular weight will predominantly be on the weight itself; effects on other weights will be nominal.

\begin{equation} \label{beta approx}
\begin{split}
\beta_i(t+1) & = \beta_i(t) - \frac{1}{2} \theta \sum_j \frac{\partial \delta^2(t)}{\partial \boldsymbol{w}_j(t)} \frac{\partial \boldsymbol{w}_j(t)}{\partial \beta_i} \\
         & \approx \beta_i(t) - \frac{1}{2} \theta \frac{\partial \delta^2(t)}{\partial \boldsymbol{w}_i(t)} \frac{\partial \boldsymbol{w}_i{(t)}}{\partial\beta_i}
\end{split}
\end{equation}

The use of the TD error in the gradient introduces some subtleties. The estimate of the TD error $\delta$ depends on the predicted value of the future state $V(\phi(s_{t+1}))$, resulting in a biased gradient. Since the error's target is dependent on the current values of the learned weight vector $\boldsymbol{w}$, it will not produce a true-gradient descent method \citep{barnard1993temporal}.

We use the semi-gradient, taking into account the impact of changing the weight vector or the estimate $V(\phi(s_t))$, but not on the target $R_t + \gamma V(\phi(s_{t+1}))$.  While less robust than other forms gradient descent, semi-gradient methods converge reliably and more quickly than true-gradient methods \citep{sutton1998reinforcement}.

\begin{equation} \label{beta derivation}
\begin{split}
\beta_i(t+1) & \approx \beta_i(t) - \theta \delta(t) \frac{\partial \delta(t)}{\partial \boldsymbol{w}_i(t)} \frac{\partial \boldsymbol{w}_i(t)}{\partial\beta_i} \\
& = \beta_i(t) - \theta \delta(t) \frac{\partial [-\boldsymbol{w}_i(t) \phi_i(t)]}{\partial \boldsymbol{w}_i(t)} \frac{\partial \boldsymbol{w}_i(t)}{\partial\beta_i} \\
& = \beta_i(t) + \theta \delta(t) \phi_i(t)\frac{\partial w_i(t)}{\partial\beta_i} \\
& =\beta_i(t) + \theta \delta(t)\phi_i(t)\boldsymbol{H}_i(t)
\end{split}
\end{equation}

We then complete the simplification of $\beta$'s update by defining an additional memory vector $\boldsymbol{H}$ as $\frac{\partial \boldsymbol{w}_i(t+1)}{\partial \beta_i}$. We then derive the update for $\boldsymbol{H}$.

\begin{equation} \label{h_partial}
\begin{split}
\boldsymbol{H}_i(t+1) & = \frac{\partial \boldsymbol{w}_i(t+1)}{\partial \beta_i} \\
	   	 & = \frac{\partial[ \boldsymbol{w}_i(t) + e^{\beta_i(t+1)} \delta(t) z_i(t)]}{\partial \beta_i} \\
         & = \boldsymbol{H}_i(t) + e^{\beta_i(t+1)}\delta(t)z_i(t)+e^{\beta_i(t+1)}\frac{\partial \delta(t)}{\partial \beta_i} z_i(t) + e^{\beta_i(t+1)}\frac{\partial z_i(t)}{\partial \beta_i} \delta_i(t)
\end{split}
\end{equation}

This simplification leaves us with $\frac{\partial \delta(t)}{\partial \beta_i}$, derived in (\ref{eq1}), and $\frac{\partial z_i(t)}{\partial \beta_i}$, derived in (\ref{z}). We use the same approximation as in (\ref{beta approx}) to simplify:

\begin{equation} \label{eq1}
\begin{split}
\frac{\partial \delta(t)}{\partial \beta_i} & =\frac{\partial}{\partial \beta_i}[-V(\phi(t))] \\
											& =\frac{\partial}{\partial \beta_i} [- \sum_j w_j(t)\phi_j(t)] \\
                                            & \approx \frac{\partial}{\partial\beta_i}[-w_i(t)\phi_i(t)] = -H_i(t) \phi_i(t)
\end{split}
\end{equation}

We finally simplify the following:

\begin{equation} \label{z}
\begin{split}
\frac{\partial z_i(t+1)}{\partial \beta_i} = \frac{\partial}{\partial \beta_i}[\gamma \lambda z_i(t) + \phi_i(t)] = \frac{\partial z_i(t)\gamma \lambda}{\partial \beta_i} = 0
\end{split}
\end{equation}

We see that (\ref{z}) results in a decaying trace of the initialized value of the eligibility traces. Since eligibility traces are initialized to 0, this value will always be 0.

\begin{equation} \label{h done}
\begin{split}
H_i(t+1) 	& \approx H_i(t) + e^{\beta_i(t+1)}\delta(t)z_i(t) - e^{\beta_i(t+1)}H_i(t) \phi_i(t) z_i(t)  + 0 e^{\beta_i(t+1)} \delta_i(t)\\ 
			& \approx H_i(t)[1-\alpha_i(t+1) \phi_i(t) z_i(t)] + \alpha_i(t+1)\delta(t)z_i(t)
\end{split}
\end{equation}

We then take the results from (\ref{eq1}) and (\ref{z}) to complete the definition of $\boldsymbol{H}$'s update. After positively bounding $[1-\alpha(t+1) \phi_i(t) z_i(t)]$, TIDBD can be seen to be a form of stochastic meta-descent for the parameter $\beta$ which updates our step-size. The update for $H$ and $\beta$ may then be implemented directly as shown earlier in Algorithm \ref{TIDBD_alg}

\section{Does TIDBD outperform ordinary TD?} 
First, we assess the ability of TIDBD to improve upon traditional TD prediction in a simple tabular setting. In a tabular setting the advantages of vectorizing step-sizes and performing representation learning are abstracted away. By using a stationary tabular problem, we assess whether adapting step-sizes with TIDBD is an improvement over ordinary TD in general. 

\begin{figure}[H]
\centering
\begin{subfigure}{0.5\textwidth}
\centering
\includegraphics[width=\linewidth,height=0.17\textheight, keepaspectratio]{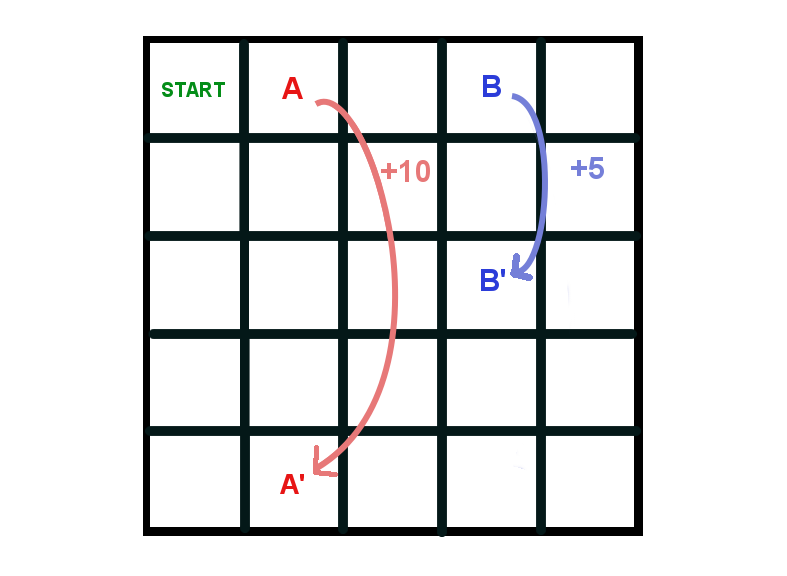}
\caption{The grid world.}
\label{dynamics}
\end{subfigure}%
\begin{subfigure}{0.5\textwidth}
    \centering
    \includegraphics[width=\linewidth,height=0.17\textheight, keepaspectratio]{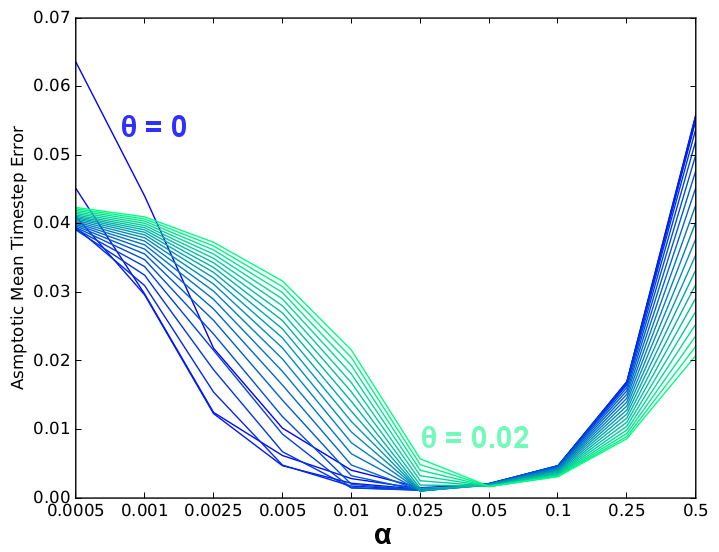}
    \caption{Asymptotic MSVPE of TIDBD(0) for all $\theta$ and $\boldsymbol{\alpha}$}
    \label{performance_exp_1_all}
\end{subfigure}
\caption{The gridworld experiment.}
\end{figure}

To make a suitable prediction task, we created a MRP from a grid-world problem originally described in \citealt{sutton1998reinforcement}. As depicted in Figure~\ref{dynamics}, each tile in the 5 $\times$ 5 grid-world represents a state. The state transitions were the four cardinal directions---north, south, east, and west---which were taken by a equiprobable random policy. A transition which leaves the grid results in the agent staying in the same state and a reward of -1. Regardless of the transition taken in A or B, the learner transitions to state A' and B' with probability 1. A transition from A to A' yields a reward of 10 and a transition from B to B' yields a reward of 5. All other transitions receive a reward of 0. The start state was the top left-hand corner. A trial consisted of each prediction method learning a value function while the equiprobable random policy moved the agent around the grid world for 15000 time-steps.

\begin{figure}[H]
    \centering
    \includegraphics[width=\textwidth]{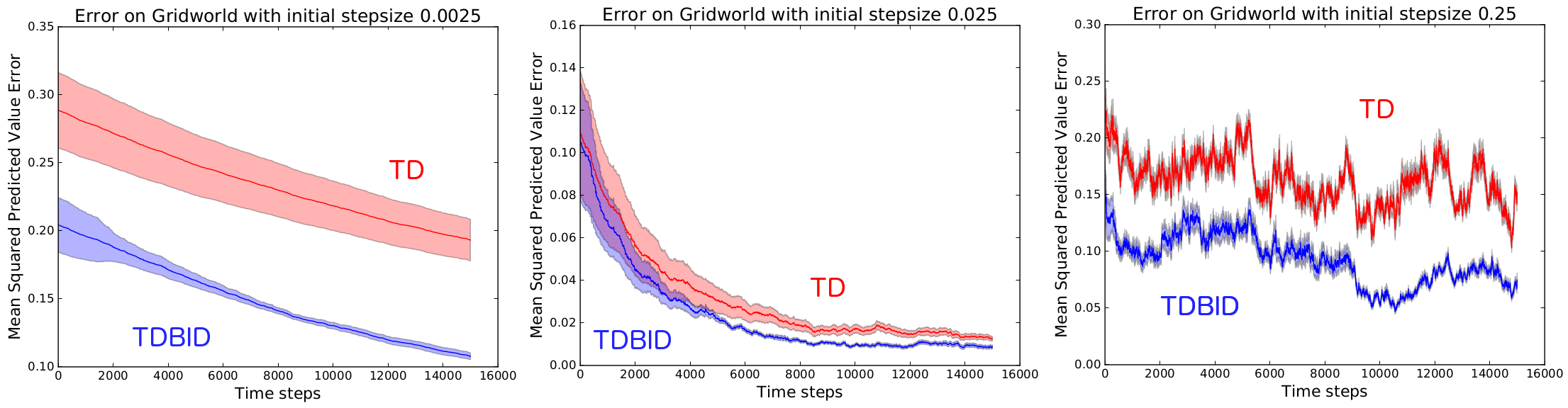}
	\caption{Comparison of TD(0) and TIDBD(0) for a variety of initial $\boldsymbol{\alpha}$ values.}
	\label{performance_exp_1}
\end{figure}

We compared TIDBD and TD on initial step-sizes distributed between 0.0005 and 0.5. For both methods $\lambda = 0$ and  $\gamma = 0.99$. We swept over a 21 different meta-parameters equally distributed within the range of $0<\theta<0.2$. Note, when $\theta = 0$, TIDBD and TD are equivalent. The mean squared value error (MSVE) was calculated at each time-step using the difference between the actual value of all states and the current learned estimate.

Figure~\ref{performance_exp_1_all} shows the performance for all values of $\theta$ at all $\alpha$ values, suggesting the sensitivity of TIDBD to $\theta$ choices in this domain. Figure~\ref{performance_exp_1} depicts the MSPVE of the TD and the TIDBD predictors averaged across 30 runs for sample initial $\alpha$s across the sweep. Although for the best initial step-size (0.0025) TIDBD has only a modest improvement over TD, it still has lower variance. For every tested initial $\alpha$, there was a corresponding $\theta$ for which TIDBD improved upon ordinary TD. 

In general, tuned TIDBD outperforms ordinary TD by adapting step-sizes online, attaining lower variance and error on a simple tabular prediction task.

\section{Can TIDBD perform representation learning?} 

\begin{wrapfigure}{r}{0.5\textwidth}
  \begin{center}
    \includegraphics[height=0.17\textheight, keepaspectratio]{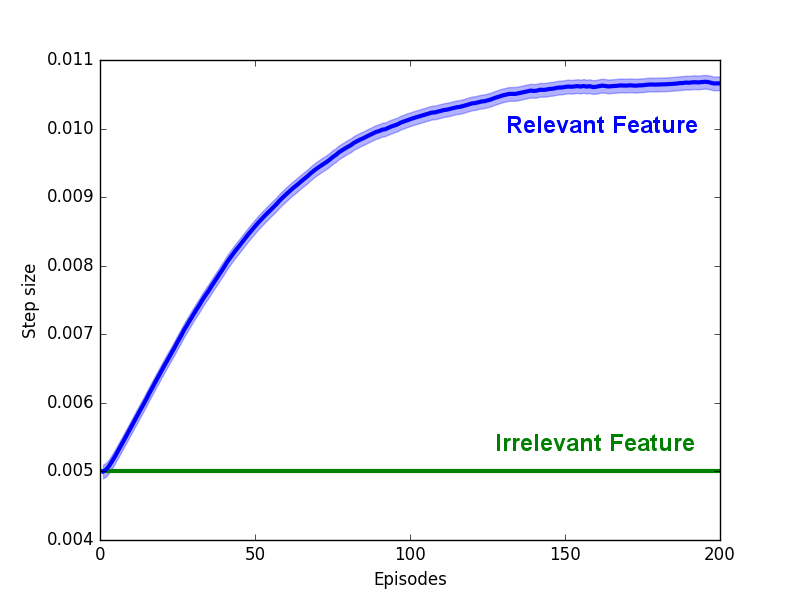}
  \end{center}
  \caption{Features' Step-sizes on mountaincar.}
    \label{stepsizeplot}
\end{wrapfigure}

In our second experiment we explore the benefits of using a vectorized step-size. In particular, we evaluate TIDBD's ability to give large step-sizes to relevant features while giving irrelevant features small step-sizes.

In a fashion similar to the grid-world experiment, we used the classic mountain-car problem to construct a MRP prediction problem. The goal of mountain-car is to reach the top of a steep hill in the shortest time possible. We trained a SARSA(0) agent with $\gamma=0.99$ to reach the top of the hill in 160 time-steps and used this policy to define the transitions in our MRP. The TIDBD learner evaluates the value of a state based on the rewards received as the SARSA agent drives the car.

For the TIDBD learner $\gamma = 0.99$ and $\lambda = 0$. Its state space consists of the real-valued position, velocity of the car, and 10 real-valued random numbers. These random values were added to act as irrelevant features, with which we asses TIDBD's ability to perform representation learning. The resulting state is tile-coded with 10 tilings of size 10~$ \times $~10. A bias feature was concatenated to the feature vector, resulting in a size of 1001. 

We expect the random features to receive small step-sizes, while the step-sizes of the relevant features grow.

Figure~\ref{stepsizeplot} depicts the magnitude of the step-sizes for two different features over time: one is relevant to the task, while the other is random noise. TIDBD clearly increases the value of the relevant step-size over time, while the irrelevant feature's step-size remains close to zero. These two features are representative of the behaviour for all the other features. As anticipated, TIDBD identifies the relevant features necessary for making an accurate prediction, and assigns them larger step-sizes. 

TIDBD is able to perform basic representation learning by assigning large step-sizes to relevant features, and smaller step-sizes to irrelevant features.

\section{How robust is TIDBD?}

In our final experiment, we determine how well TIDBD performs relative to both ordinary TD and other adaptive step-size methods. We examine the performance of TIDBD in comparison to ordinary TD, AlphaBound, and TD with RMSprop on a real-world robot prediction task. We assess the sensitivity of TIDBD to it's settings in comparison to other methods, and whether it can improve predictions such as those used in robot control systems.

We replicated the experiment used by \cite{van2014true} with the dataset from \cite{edwards2016application}: the learning system predicts the position of a robot gripper as a user performs a dexterity training task while in control of the robot. Such predictions can be used to build more complex control systems which adapt to the user controlling the robot \citep{edwards2016application,sherstan2015collaborative}, making it possible to continuously improve the control of these systems \citep{sherstan2015collaborative}. A great challenge for forming these predictions is finding and setting appropriate parameters for a given prediction method. End-user time is precious, and designers cannot possibly test their prediction algorithms on datasets which are representative of all the situations the robot might encounter.

We performed a sweep over parameter settings in order to compare the TD methods for a selection of $\lambda$ values between 0 and 1. For all methods $\gamma = 0.97$. Ordinary TD methods were swept over $\alpha$ settings between 0 and 2. AlphaBound was initialized with an initial step-size of 1, as specified in \cite{dabney2012adaptive}. The initial step-size for both TD RMSprop and TIDBD was set to $\frac{0.5}{9}$, where 9 is the number of active features. We are assessing whether or not TIDBD is easier to tune than other methods, so we choose an intuitively good initial step-size, but not necessarily the the best possible value. To make RMSprop suitable for TD methods, we use the semi-gradient in its calculated weighted average. For TD RMSprop, $\epsilon = 10^{-8}$, an decay rates were varied between 0 and 1. For TIDBID, $\theta$ values were swept between 0 an 0.02. 

Our error is the difference between the true return of the hand position and the predicted position at each time-step. Algorithm performance is assessed at each combination of parameter settings using the cumulative absolute prediction error averaged over 24 independent trials of user data.

\begin{figure}[h]
\centering
\includegraphics[keepaspectratio, height=0.25\textheight]{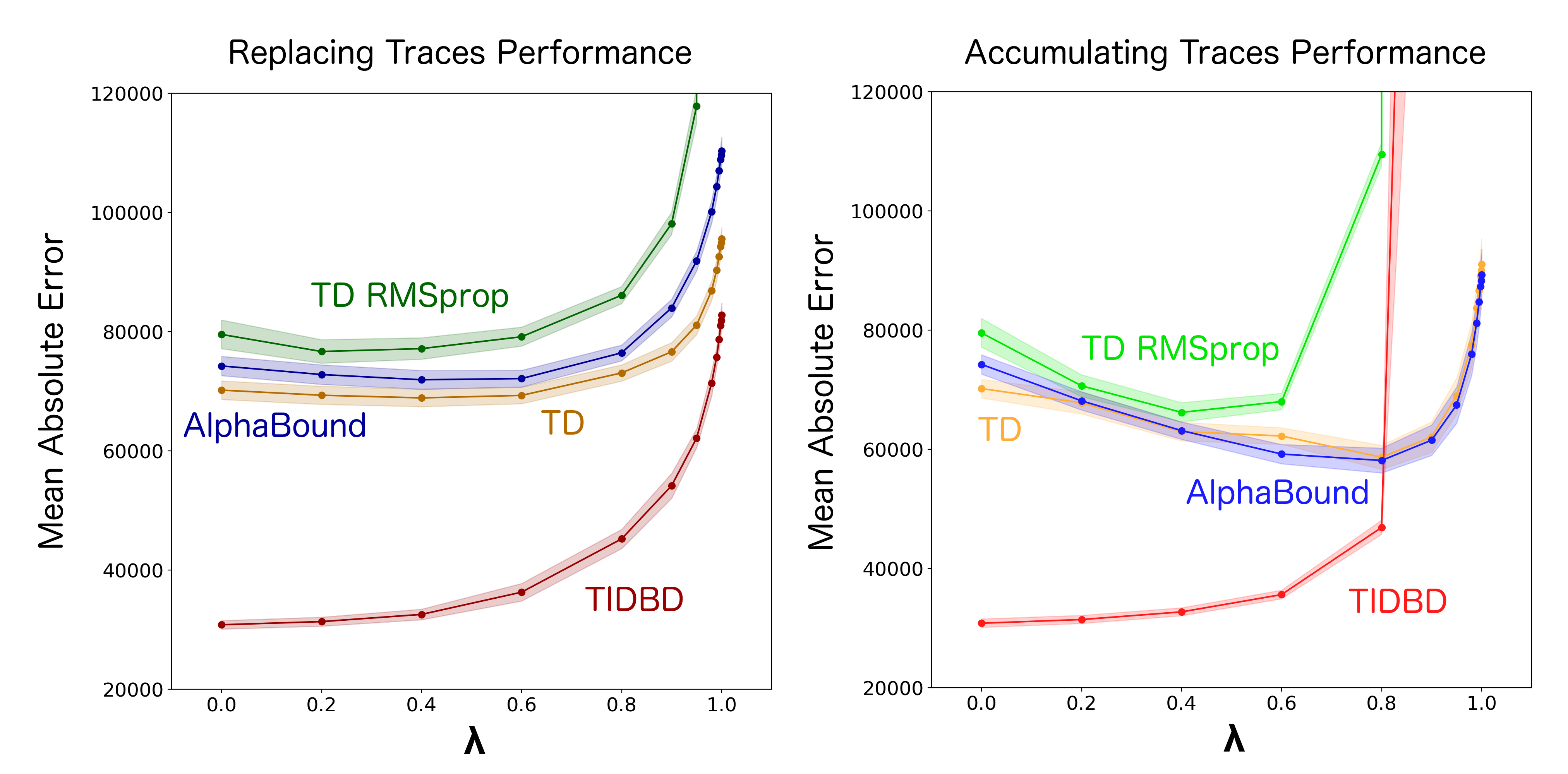}
\centering
\caption{Comparison of TIDBD to other learning methods on the robotic prediction task.}
\label{sweep_perf}
\end{figure}

Each point in Figure~\ref{sweep_perf} represents the best performing parameter settings for each $\lambda$ value. TIDBD with replacing traces outperforms all other methods at every $\lambda$, especially for low values of $\lambda$ In addition, for both variants of traces, TIDBD has low variance in comparison to other methods. Interestingly, TIDBD with accumulating traces suffers at higher $\lambda$ settings. The origin of this performance drop is in the the interaction between our adaptive $\boldsymbol{\alpha}$ and our eligibility traces. Accumulating traces will continue to assign credit to states if we visit them multiple times. If a state has been visited multiple times before the weighting has decayed from the eligibility, the weight update $\boldsymbol{\alpha} \delta \boldsymbol{z}$ will have a greater weight due to $\boldsymbol{z}$ assigning greater credit. Similarly, our TIDBD uses a correlation of recent weight updates to determine the step size of a feature. If a weight's updates are highly correlated, it's learning rate will adjust according. Together, these interactions can cause instability for larger meta-step-sizes. 

\begin{figure}[htbp!]
    \centering
    \includegraphics[width=\textwidth,height=0.4\textheight, keepaspectratio]{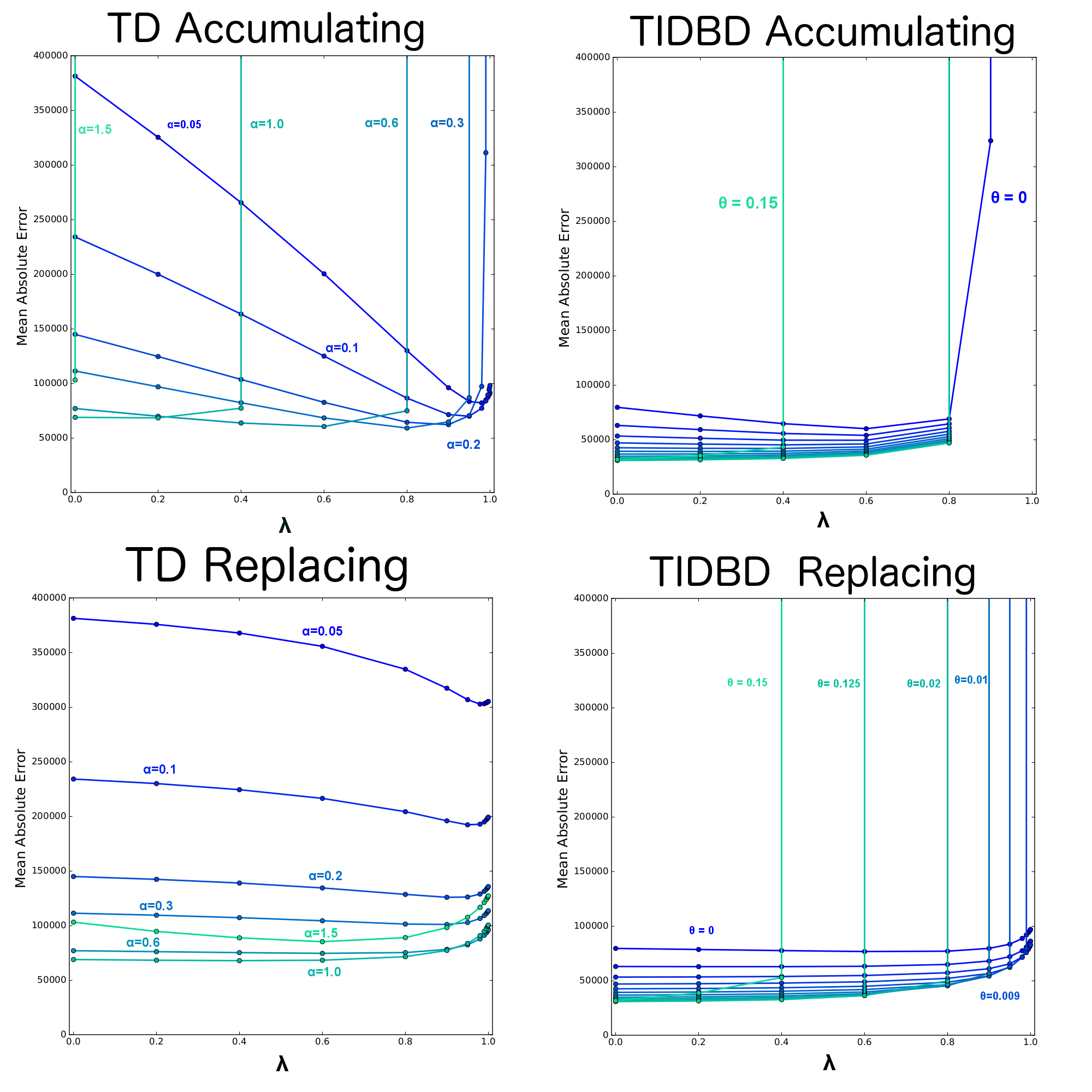}  
    \caption{Error on prediction task for all $\alpha$ and $\lambda$ combinations in the sweep.}
  \label{comparison_step} 
\end{figure}

In Figure \ref{comparison_step} we show a parameter study which demonstrates the parameter sensitivity of ordinary TD and TIDBD. Each sub-figure shows the error at each setting of $\lambda$ for all $\boldsymbol{\alpha}$ or $\theta$ values in our given sweep. When observing the sensitivity of ordinary TD, we can see that $\alpha$ values similar to the initial $\alpha$ we used for TIDBD perform as expected: they perform relatively well, but are not the best settings. Despite the use of a sub-optimal initial $\alpha$, TIDBD was able to outperform the ordinary TD methods for most $\theta$ values.

In Figure \ref{comparison_step} we see that the performance of TIDBD in comparison to meta-step-size $\theta$'s setting is much less variable than the ordinary TD counterparts. The sensitivity of TIDBD with accumulating traces further elaborates on performance problems with higher $\lambda$ values, depicting how with larger meta-step-sizes and larger $\lambda$ values divergence can occur. Of note is TIDBD's performance with replacing traces. We can see that even when TIDBD uses a sub-optimal setting for the ordinary TD methods, any of the meta-step-sizes in our sweep brings the performance to the same level of the best tuned TD methods, or even better. However, this comes at risk of instability.

\section{Conclusion, limitations, and future work}

We presented an approach of generalizing Delta-Bar-Delta to temporal-difference learning and demonstrate that the effectiveness of IDBD carries over from supervised learning to TD. We derived TIDBD as stochastic meta-descent over the step-size parameter. We demonstrated that adapting step-sizes with TIDBD improves over ordinary TD methods with tuned static step-sizes, even on stationary problems. On non-stationary tasks, we showed that TIDBD is able to find appropriate step-sizes in general; TIDBD can discriminate between relevant and irrelevant features, giving appropriate step-sizes to each. 

We examined TIDBD's performance against both ordinary TD and other adaptive methods on data from a real-world robotic prediction problem. We found that for an ordinary, intuitive step-size which is commonly used on such tasks, the sensitivity to TIDBD's meta-step-size was far less than the sensitivity to ordinary TD's step-size. Using a sub-optimal step-size and a tuned meta-parameter, TIDBD outperformed ordinary TD and had less variance in average error. In addition, TIDBD outperformed the scalar step-size adaptation method AlphaBound for each parameter setting

The greatest limitation of TIDBD is the need to tune the meta-parameter. While a limitation, there are signs that TIDBD is less sensitive to $\theta$ than TD is to $\alpha$. In addition this sensitivity could be possibly be mitigated by further extending TIDBD to include AutoStep's \citep{mahmood2012tuning} normalization. This would bring greater stability to TIDBD by attempting to eliminate divergence when using adaptive step-sizes. Additional stability can be sought out by finding ways of better integrating traces into TIDBD to prevent over-compensation in applications where $\lambda$ is close to 1.0. This work focused solely on prediction tasks as a testbed to assess generalization: in the future, it would be desirable to assess TIDBD's performance on control tasks.

In summary, we generalized IDBD to TD, creating TIDBD: an adaptive step-size method for TD learning which satisfies our requirement of being vectorized, and enabling step-sizes to increase and decrease in value. TIDBD shows promise for adapting step-sizes and performing representation learning for TD learning, enabling better solutions for life-long continual learning problems.

\end{document}